\documentclass[12pt, a4paper, onecolumn]{article}

\usepackage{fullpage}
\usepackage{hyperref}

\usepackage[english]{babel}
\usepackage{amsmath}         
\usepackage{amssymb}         

\usepackage{graphicx}
\usepackage{url}           

\long\def\symbolfootnote[#1]#2{
\begingroup
	\def\thefootnote{\fnsymbol{footnote}}\footnote[#1]{#2}
\endgroup}

\newcommand{\DuARTE}[0]{\textsc{Edgar}}
\newcommand{\DuARTEm}[0]{\textsc{Edgar }}
\newcommand{\JaTeDigo}[0]{\textsc{JaTeDigo}}
\newcommand{\JaTeDigom}[0]{\textsc{JaTeDigo }}

\begin{document}

\title{Towards the Rapid Development of a Natural Language Understanding Module}

\author{Catarina Moreira\\ \small \texttt{catarina.p.moreira@ist.utl.pt}\\
\and
Ana Cristina Mendes\\ \small \texttt{ana.mendes@l2f.inesc-id.pt}
\and
Lu\'{i}sa Coheur\\ \small \texttt{luisa.coheur@l2f.inesc-id.pt}
\and
Bruno Martins\\ \small \texttt{bruno.g.martins@ist.utl.pt}
\and
\\Instituto Superior T\'{e}cnico, INESC-ID\\ Av. Professor Cavaco Silva, 2744-016 Porto Salvo, Portugal\\  
\\ \small The original publication is available at: Intelligent Virtual Agents, Springer\\
 \small \text{\url{http://link.springer.com/chapter/10.1007\%2F978-3-642-23974-8\_33}}
}

\date{}

\maketitle              

\begin{abstract}
When developing a conversational agent, there is often an urgent need to have a prototype available in order to test the application with real users. A Wizard of Oz is a possibility, but sometimes the agent should be simply deployed in the environment where it will be used. Here, the agent should be able to capture as many interactions as possible and to understand how people react to failure. In this paper, we focus on the rapid development of a natural language understanding module by non experts. Our approach follows the learning paradigm and sees the process of understanding natural language as a classification problem. 
We test our module with a conversational agent that answers questions in the art domain. Moreover, we show how our approach can be used by a natural language interface to a cinema database.
\end{abstract}

\section{Introduction}
In order to have a clear notion of how people interact with a conversational agent, ideally the agent should be deployed at its final location, so that it can be used by people sharing the characteristics of the final users. This scenario allows the developers of the agent to collect corpora of real interactions. Although the Wizard of Oz technique~\cite{Kelley84natural} can also provide these corpora, sometimes it is not a solution if one needs to test the system with many different real users during a long period and/or it is not predictable when the users will be available.

The natural language understanding (NLU) module is one of the most important components in a conversational agent, responsible for interpreting the user requests. The symbolic approach to NLU usually involves a certain level of natural language processing, which includes hand crafted grammars and requires a certain amount of expertise to develop them; by the same token, the statistical approach relies on a large quantity of labeled corpora, which is often not available. 

\begin{figure}[h!]
\resizebox{\columnwidth}{!} {
\includegraphics[scale = 0.5]{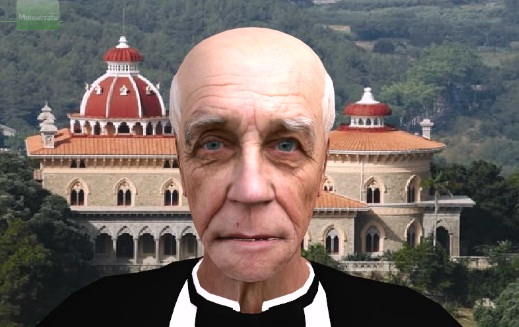}
}
\caption{Agent Edgar}
\label{fig:edgar}
\end{figure}

In this paper we hypothesize that a very simple and yet effective NLU module can be built if we model the process of NLU as a classification problem, within the machine learning paradigm. Here, we follow the approach described in \cite{Bhagat}, although their focus is on frame-based dialogue systems.  
Our approach is language independent and does not impose any level of expertise to the developer: he/she simply has to provide the module with a set of possible interactions (the only constraint being the input format) and a dictionary (if needed). Given this input, each interaction is automatically associated with a virtual category and a classification model is learned. The model will map future interactions in the appropriate semantic representation, which can be a logical form, a frame, a sentence, etc.
We test our approach in the development of a NLU module for \DuARTE (Figure ~\ref{fig:edgar}) a conversational agent operating in the art domain. Also, we show how the approach can be successfully used to create a NLU module for a natural language interface to a cinema database, \JaTeDigo, responsible for mapping the user requests into logical forms that will afterwards be mapped into SQL queries\footnote{All the code used in this work will be made available for research purposes at \\ \url{http://qa.l2f.inesc-id.pt/}.}.

The paper is organized as follows: in Section \ref{RW} we present some related work and in Section \ref{NLU} we describe our NLU module. Finally, in Section \ref{exp} we show our experiments and in Section \ref{conc} we conclude and present future work directions.

\section{Related Work}\label{RW}

NLU is the task of mapping natural language utterances into structures that the machine can deal with: the semantic representation of the utterances. The semantics of a utterance can be a logical form, a frame or a natural language sentence already understood by the machine.
The techniques for NLU can be roughly split into two categories: symbolic and sub-symbolic. There are also hybrid techniques, that use characteristics of both categories.

Regarding symbolic NLU, it includes keyword detection, pattern matching and rule-based techniques. For instance, the virtual therapist ELIZA \cite{Weizenbaum66} is a classical example of a system based on pattern matching. Many early systems were based on a sophisticated syntax/semantics interface, where each syntactic rule is associated with a semantic rule and logical forms are generated in a bottom-up, compositional process. Variations of this approach are described in \cite{Allen,Jurafsky}. Recently, many systems follow the symbolic approach, by using in-house rule-based NLU modules \cite{HCA,max}. However, some systems use the NLU modules of available dialogue frameworks, like the Let's Go system \cite{LetsGO}, which uses Olympus\footnote{\url{http://wiki.speech.cs.cmu.edu/olympus/index.php/Olympus.}}.

In what concerns sub-symbolic NLU, some systems receive text as input \cite{Bhagat} and many are dealing with transcriptions from an Automatic Speech Recognizer \cite{Statistical}. In fact, considering speech understanding, the new trends considers NLU from a machine learning point of view. However, such systems usually need large quantities of labeled data and, in addition, training requires a previous matching of words into their semantic meanings.

\section{The natural language understanding module}\label{NLU}

\begin{figure}[ht]
\resizebox{\columnwidth}{!} {
\includegraphics[width=1\textwidth]{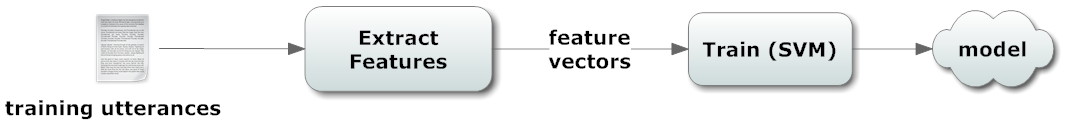}
}
\caption{Training the NLU module.}
\label{fig:train}
\end{figure}

The NLU module receives as input a file with possible interactions (the training utterances file), from which several features are extracted. These features are in turn used as input to a classifier. In our implementation, we have used 
Support Vector Machines (SVM) as the classifier and the features are unigrams. However, in
order to refine the results, other features can easily be included. Figure \ref{fig:train} describes the training phase of the NLU module.\\

Each interaction specified in the training utterances file is a pair, where the first element is a set of utterances that paraphrase each other and that will trigger the same response; the second element is a set of answers that represent possible responses to the previous utterances. That is, each utterance in one interaction represents different manners of expressing the same thing and each answer represents a possible answer to be returned by the system. The DTD of this file is the following:
\begin{verbatim}
<!ELEMENT corpus (interaction+)>
<!ELEMENT interaction (uterances, answers)>
<!ELEMENT utterances (u+)>
<!ELEMENT answers (a+)>
<!ELEMENT u (#PCDATA)>
<!ELEMENT a (#PCDATA)>  
\end{verbatim}

The NLU module also accepts as input a dictionary, containing elements to be replaced with labels that represent broader categories. Thus, and considering that \textsc{tag} is the label that replaces a compound term w$_{1}$... w$_{n}$ during training, the dictionary is composed of entrances in the format:\\ 

\textsc{tag} w$_{1}$... w$_{n}$~~~ (for example: \textsc{actor} Robert de Niro) \\

If the dictionary is used, Named Entity Recognition (NER) is performed to replace the terms that occur both in the training utterances file and user utterances. This process uses the LingPipe\footnote{\url{http://alias-i.com/lingpipe/}.} implementation of the Aho-Corasick algorithm \cite{Aho}, that searches for matches against a dictionary in linear time in terms of the length of the text, independently of the size of the dictionary.

A unique identifier is then given to every paraphrase in each interaction -- the interaction category -- which will be the target of the training. For instance, since sentences \textit{H\'{a} alguma data prevista para a conclus\~{a}o das obras?} and \textit{As obras v\~{a}o acabar quando?} ask for the same information (\textit{When will the conservation works finish?}), they are both labeled with the same category, generated during training: \texttt{agent\_7}. The resulting file is afterwards used to train the classifier. 

\begin{figure}[h!t]
\resizebox{\columnwidth}{!} {
\includegraphics[width=1\textwidth]{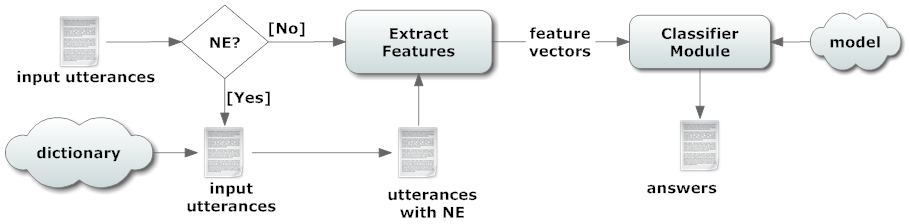}
}
\caption{Pipeline of the NLU module.}
\label{fig:default}

\end{figure}

After the training phase, the NLU module receives as input a user utterance. If the NE flag is enabled, there is a pre-processing stage, where the NE recognizer tags the named entities in the user utterance before sending it to the classifier. Then the classifier chooses a category for the utterance. Since each category is associated with a specific interaction (and with its respective answers), one answer is randomly chosen and returned to the user. These answers must be provided in a file with the format \texttt{category answer}. Notice that more than one answer can be specified. Figure \ref{fig:default} describes the general pipeline of the NLU module. 

\section{Experiments}\label{exp}

This section presents the validation methodology and the obtained results.

\subsection{Experimental setup}

In order to test our approach to the rapid development of a NLU module, we first collected a corpus that contains interactions in the art domain: the \textsc{Art} corpus. It was built to train \DuARTE, a conversational agent whose task is to engage in inquiry-oriented conversations with users, teaching about the Monserrate Palace. \DuARTEm answers questions on its domain of knowledge, although it also responds to questions about himself. The \textsc{Art} corpus has 283 utterances with 1471 words, from which 279 are unique. The utterances represent 52 different interactions (thus, having each interaction an average of 5.4 paraphrases).

For our experiments in the cinema domain, we have used the \textsc{Cinema} corpus, containing 229 questions mapped into 28 different logical forms, each one representing different SQL queries. A dictionary was also build containing actor names and movie titles.



\subsection{Results}

The focus of the first experiment was to chose a correct answer to a given utterance. This scenario implies the correct association of the utterance to the set of its paraphrases. For instance, considering the previous example sentence \textit{As obras v\~{a}o acabar quando?}, it should be associated to the category \texttt{agent\_7} (the category of its paraphrases). 

The focus of the second experiment was to map a question into an intermediate representation language (a logical form) \cite{androutsopoulos95natural}. For instance, sentence  \textit{Que actriz contracena com Viggo Mortensen no Senhor dos An\'{e}is?} (\textit{Which actress plays with Viggo Mortensen in The Lord of the Rings?}) should be mapped into the form \texttt{WHO\_ACTS\_WITH\_IN(Viggo Mortensen,  The Lord of the Rings)}.


Both corpora where randomly split in two parts (70\%/30\%), being 70\% used for training and 30\% for testing. This process was repeated 5 times. Results are shown in Table \ref{table:edgar}.

\begin{table}[t!h!]
\resizebox{\columnwidth}{!} {
\begin{tabular}{|c||c|c|c|c|c||c|}
\hline
~ Corpus  ~& ~ fold 1 ~ & ~ fold 2 ~ & ~ fold 3 ~  & ~ fold 4 ~ & ~ fold 5 ~ & ~ average ~\\
\hline
\textsc{Art} &  0.78 & 0.74 & 0.86 & 0.87 & 0.92 & 0.83\\
\hline
\textsc{Cinema} & 0.87 & 0.90 & 0.79 & 0.77 & 0.82 & 0.83\\
\hline
\end{tabular}
}
\caption{Accuracy results}
\label{table:edgar}
\end{table}

\subsection{Discussion}

From the analysis of Table~\ref{table:edgar}, we conclude that a simple technique can lead to very interesting results. Specially if we compare the accuracy obtained for the \textsc{Cinema} corpus with previous results of 75\%, which were achieved with recourse to a linguistically rich framework that required several months of skilled labour to build. Indeed, the previous implementation of \JaTeDigom was based on a natural language processing chain, responsible for a morpho-syntactic analysis, named entity recognition and rule-based semantic interpretation.

Another conclusion is that one can easily develop an NLU module. In less than one hour we can have the set of interactions needed for training and, from there, the creation of the NLU module for that domain is straightforward. Moreover, new information can be easily added, allowing to retrain the model.

Nevertheless, we are aware of the debilities of our approach. The NLU module is highly dependent of the words used during training and the detection of paraphrases is only successful for utterances that share many words. In addition, as we are just using unigrams as features, no word is being detached within the input utterances, resulting in some errors. For instance, in the second experiment, the sentence \textit{Qual o elenco do filme MOVIE?} (\textit{Who is part of MOVIE's cast?}) was wrongly mapped into \texttt{QT\_WHO\-\_MAIN\-\_ACT(MOVIE)}, although very similar sentences existed in the training. A solution for this problem is to add extra weight to some words, something that could be easily added as a feature if these words were identified in a list. Moreover, adding synonyms to the training utterances file could also help.

Another limitation is that the actual model does not comprise any history of the interactions. Also, we should carefully analyze the behavior of the system with the growing of the number of interactions (or logical forms), as the classification process becomes more complex.

\section{Conclusions and Future Work}\label{conc}

We have presented an approach for the rapid development of a NLU module based on a set of possible interactions. This approach treats the natural language understanding problem as a classification process, where utterances that are paraphrases of each other are given the same category. It receives as input two files, the only constraint being to write them in a given xml format, making it very simple to use, even by non-experts. Moreover, it obtains very promising results. 
As future work, and although moving from the language independence, we would like to experiment additional features and we would also like to try to automatically enrich the dictionary and the training files with relations extracted from WordNet.

\section*{Acknowledgments}

This work was supported by FCT (INESC-ID multiannual funding) through the PIDDAC Program funds, and also through the project FALACOMIGO (Projecto em co-promo\c{c}\~{a}o, QREN nº 13449). Ana Cristina Mendes is supported by a PhD fellowship from Funda\c{c}\~{a}o para a Ci\^{e}ncia e a Tecnologia (SFRH/BD/43487/2008).
%
%
%

\bibliographystyle{plain}

\end{document}